\documentclass[11pt,letterpaper,logo]{thuair}

\usepackage[utf8]{inputenc} 
\usepackage[T1]{fontenc}    
\usepackage{hyperref}       
\usepackage{url}            
\usepackage{booktabs}       
\usepackage{amsfonts}       
\usepackage{nicefrac}       
\usepackage{microtype}      
\usepackage{xcolor}         
\usepackage[utf8]{inputenc} 
\usepackage[T1]{fontenc}    
\usepackage{hyperref}       
\usepackage{url}            
\usepackage{booktabs}       
\usepackage{amsfonts}       
\usepackage{nicefrac}       
\usepackage{microtype}      
\usepackage{xcolor}         
\usepackage{graphicx}
\usepackage{eso-pic}
\newcommand{\shijie}[1]{}

\usepackage{multirow}
\usepackage{amsmath}
\usepackage[table]{xcolor} 
\usepackage{colortbl}      
\usepackage[ruled,vlined]{algorithm2e}
\usepackage{svg}

\usepackage{graphicx}
\usepackage{wrapfig}
\usepackage{subcaption}
\usepackage{threeparttable}
\usepackage{wrapfig} 
\usepackage{graphicx}
\usepackage{fontawesome5}

\fancypagestyle{firststyle}{%
  \fancyhf{}%
  \fancyhead[L]{%
    \ifthenelse{\boolean{logo}}{%
      \includegraphics[width=200pt]{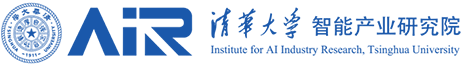}%
    }{}%
  }%
  \fancyhead[R]{%
    \ifdefined\paperurl
      \if\relax\the\paperurl\relax\else
        \href{\the\paperurl}{\urlheaderfont \itshape \the\paperurl}\\%
      \fi
    \fi
    {\footerfont\itshape\monthyeardate\today}%
  }%
  \fancyhead[C]{}%
  \fancyfoot[L]{%
    \footerfont
    \begin{minipage}[t]{0.98\textwidth}
      \textsuperscript{*}Ruofei Ju and Xinrui Wang contributed equally to this work.\\
      \textsuperscript{\dag}This work was done while Ruofei Ju and Xinrui Wang were interns at the Institute for AI Industry Research (AIR), Tsinghua University.\\
      \textsuperscript{\ddag}Corresponding authors: Xin Ding (xinding64@mail.ustc.edu.cn) and Ting Cao (tingcao@mail.tsinghua.edu.cn).
    \end{minipage}%
  }%
  \fancyfoot[C]{}%
  \fancyfoot[R]{}%
}

\title{EmbodiSkill: Skill-Aware Reflection for Self-Evolving Embodied Agents}

\author{%
  \textbf{Ruofei Ju}\textsuperscript{1}\textsuperscript{$*$}\textsuperscript{$\dag$}
  \quad
  \textbf{Xinrui Wang}\textsuperscript{2}\textsuperscript{$*$}\textsuperscript{$\dag$}
  \quad
  \textbf{Xin Ding}\textsuperscript{3}\textsuperscript{$\ddag$}
  \quad
  \textbf{Yifan Yang}\textsuperscript{4}
  \quad
  \textbf{Hao Wu}\textsuperscript{1}\hspace{4pt}
  \quad
  \textbf{Shiqi Jiang}\textsuperscript{4}
  \quad
  \textbf{Qianxi Zhang}\textsuperscript{4}\\
  \quad
  \textbf{Hao Wen}\textsuperscript{5}
  \quad
  \textbf{Xiangyu Li}\textsuperscript{5}
  \quad
  \textbf{Weijun Wang}\textsuperscript{5}
  \quad
  \textbf{Kun Li}\textsuperscript{5}
  \quad
  \textbf{Yunxin Liu}\textsuperscript{5}
  \quad
  \textbf{Haipeng Dai}\textsuperscript{1}
  \quad
  \textbf{Wei Wang}\textsuperscript{1}
  \quad
  \textbf{ Ting Cao}\textsuperscript{5}\textsuperscript{$\ddag$}

  \textsuperscript{1} Nanjing University
  \quad
  \textsuperscript{2} Huazhong University of Science and Technology 
  \quad
  \textsuperscript{3} University of Science and Technology of China \\
  \quad
  \textsuperscript{4} Microsoft Research 
  \quad 
  \textsuperscript{5} Institute for AI Industry Research (AIR), Tsinghua University
  \\
}

\begin{document}

\begin{abstract}
Embodied agents can benefit from skills that guide object search, action execution, and state changes across diverse environments. Since embodied environments vary across layouts, object states, and other execution factors, these skills must self-evolve from trajectories generated during task execution. However, existing skill self-evolution methods are mainly developed in digital environments and often convert trajectories into coarse skill updates. Directly applying this paradigm to embodied settings is problematic, because a failed task execution may reflect not only incorrect skill content, but also an execution lapse in which the agent fails to follow valid guidance. We propose \textbf{EmbodiSkill}, a training-free framework for embodied skill self-evolution through skill-aware reflection and targeted revision. EmbodiSkill interprets each trajectory with respect to the current skill, uses skill-changing evidence to update the skill body, and uses execution-lapse evidence to preserve and emphasize valid guidance. Experiments on ALFWorld and EmbodiedBench show that EmbodiSkill consistently improves embodied task success. On ALFWorld, EmbodiSkill enables a frozen Qwen3.5-27B executor to reach 93.28\% task success, outperforming GPT-5.2 used as a direct agent without skills by 31.58\%. These results show that skill-aware self-evolution helps embodied agents accumulate reusable procedural knowledge from their own trajectories. The code and data are available at
\url{https://github.com/air-embodied-brain/EmbodiSkill}.
\end{abstract}


\maketitle

\section{Introduction}

\begin{figure}[t]
    \centering
    \includegraphics[
      width=0.9\linewidth,
      height=0.62\textheight,
      keepaspectratio
    ]{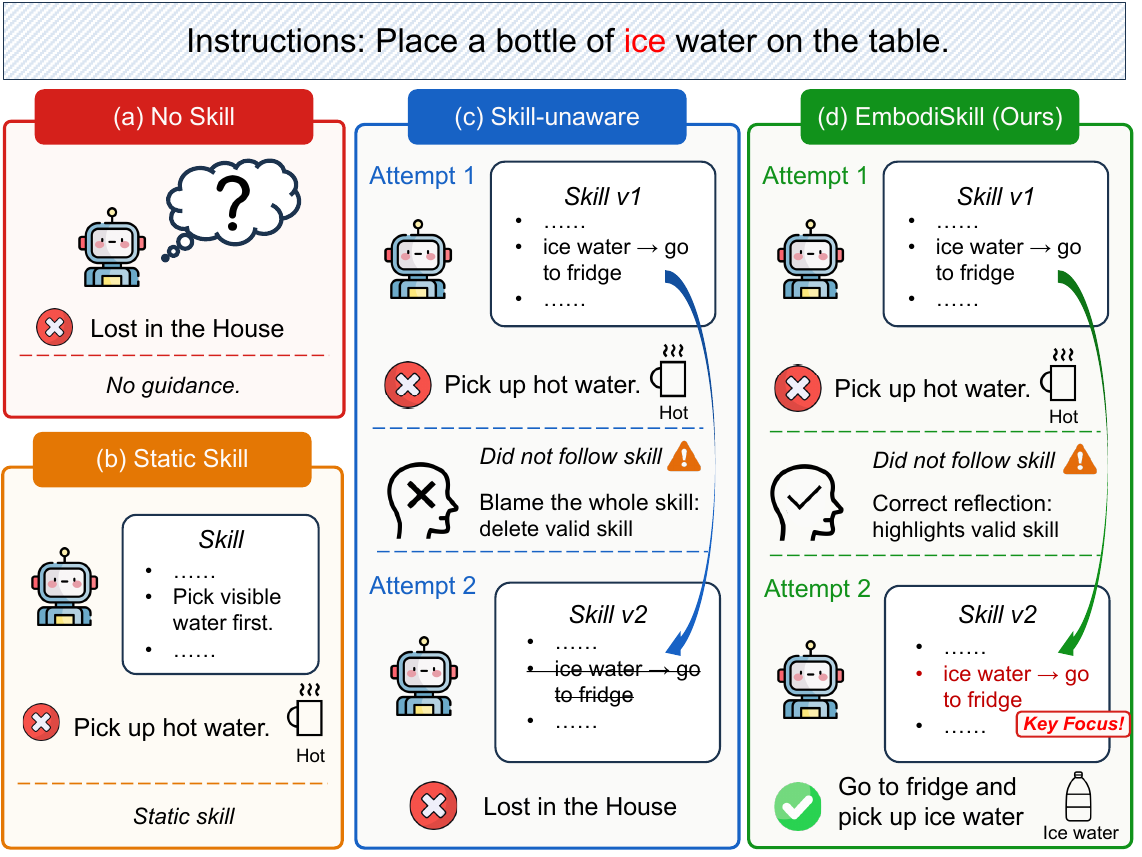}
    \caption{Motivating example of EmbodiSkill. For the same task instruction, (a) no skill leaves the agent without procedural guidance and causes inefficient exploration, (b) a static skill provides incomplete guidance and cannot adapt from trajectories, and (c) skill-unaware evolution may misinterpret an execution lapse as a skill defect and wrongly revise valid skill content. In contrast, (d) EmbodiSkill reflects on the trajectory against the current skill, preserves valid guidance, and updates the skill to improve later task executions.}
    \label{fig1}
\end{figure}

Embodied agents are expected to complete household-like tasks in physical or physically grounded 3D environments, where they must observe the scene, navigate through space, interact with objects, satisfy action preconditions, and handle failed actions~\cite{shridhar2020alfred,shridhar2021alfworld,yang2025embodiedbench}. In such tasks, the trajectories generated by different task executions often exhibit recurring procedural structures: an agent may need to locate an object before manipulating it, check whether a container is open before placing it, or adjust its viewpoint when the target is not visible. Following prior work that treats skills, action primitives, or programmatic policies as reusable procedural units for grounding high-level instructions into executable behavior~\cite{ichter2023saycan,liang2023codeaspolicies,singh2023progprompt,wang2024voyager}, we define an \textit{embodied skill} as a persistent and revisable procedural specification that guides an embodied agent across task executions. In embodied tasks, a skill specifies reusable guidance such as prerequisites, subgoal ordering, object affordances, visual-search strategies, action preconditions, and recovery strategies. Since embodied environments vary in layouts, object states, visibility conditions, and feasible action sequences, an initial skill cannot cover all situations the agent may encounter. Therefore, the central challenge is not only how an embodied agent uses a skill, but how the skill can self-evolve from trajectories into a more complete, accurate, and executable form.

Recent skill self-evolution methods show that agents can improve by extracting, revising, and reusing skills from trajectories~\cite{wang2024voyager,fang2026memp,cao2026reme,ni2026trace2skill,zhang2026coevoskills,mi2026skillpro}. However, many existing methods update skills at a coarse granularity: a trajectory is summarized into feedback, and the resulting feedback is used to coarsely update the whole skill, without explicitly identifying which skill content is implicated, why it should be changed, and how it should be changed. This update strategy becomes problematic in embodied environments, where a trajectory outcome is entangled with perception, spatial grounding, object states, action preconditions, and execution reliability~\cite{shridhar2020alfred,shridhar2021alfworld,yang2025embodiedbench,ichter2023saycan,huang2022zeroshotplanners}. A successful trajectory may reveal useful new skill content or a better way to perform an existing skill, but a failed trajectory does not necessarily mean that the current skill is wrong. It may instead result from the agent failing to follow valid skill content~\cite{ichter2023saycan,huang2022zeroshotplanners,shridhar2021alfworld}. If such trajectories are converted into coarse skill updates, the evolving skill may accumulate redundant content, overwrite valid guidance, or preserve incorrect prescriptions. Figure~\ref{fig1} illustrates this problem with a motivating example, where skill-unaware evolution misinterprets a failed task execution and incorrectly revises valid skill content.

To address these problems, we propose EmbodiSkill, a framework for embodied skill self-evolution through skill-aware reflection and a skill-aware evolution spiral. Given a trajectory and the current skill, EmbodiSkill does not convert the trajectory into general feedback for rewriting the whole skill. Instead, skill-aware reflection uses the trajectory to examine the current skill and determine which skill content should be added, optimized, corrected, or preserved. A trajectory can lead to DISCOVERY when it reveals missing skill content, OPTIMIZATION when it suggests a better way to perform an existing skill, SKILL DEFECT when it exposes incorrect, incomplete, or underspecified skill content, or EXECUTION LAPSE when the skill is valid but the agent fails to execute it. The resulting reflections serve as targeted update signals for skill revision. In the skill-aware evolution spiral, justified reflections are integrated into the next skill version, and the revised skill is then used to guide subsequent task executions that generate new trajectories. This creates a closed loop in which trajectories refine the skill, and the refined skill guides later task executions that produce further trajectories for skill evolution. By separating these cases, EmbodiSkill revises only the skill content that should change while preserving valid skill content from unnecessary updates. This selective revision process allows the skill to evolve progressively: each trajectory contributes targeted evidence, and the skill becomes increasingly complete, accurate, and executable over time.

We evaluate EmbodiSkill on embodied and interactive task benchmarks covering visual object interaction, navigation, and household task completion~\cite{shridhar2021alfworld,yang2025embodiedbench}. On ALFWorld, EmbodiSkill enables a local open-weight Qwen3.5-27B executor to achieve 93.28\% task success, outperforming GPT-5.2 used as a direct agent by 31.58\%. It also exceeds the representative memory-based baseline G-Memory~\cite{zhang2025gmemory} by 25.01\%. More importantly, compared with the skill-unaware variant using the same skill evolution model, EmbodiSkill brings a 19.04\% relative improvement, demonstrating the importance of skill-aware reflection. These results show that selectively updating the implicated skill content is more effective than coarsely revising the skill, allowing the skill to become progressively more complete, accurate, and executable.




\section{Related Work}
\label{sec:related_work}

\subsection{Memory-Based Methods for Embodied Agents}

In embodied environments, agents often face recurring objects, spatial layouts, action constraints, and failure patterns across task executions, which are reflected in the resulting trajectories. A common way to exploit such recurrence is to adapt memory-based agent methods, which store past trajectories, summaries, or structured notes, and retrieve relevant memory to guide later decision making. For example, Reflexion stores verbal feedback from previous trials as episodic memory~\cite{shinn2023reflexion}, while ExpeL extracts reusable lessons from successful and failed trajectories~\cite{zhao2024expel}. Recent memory systems such as Mem0, G-Memory, A-MEM, and LangMem further organize agent memory for long-term reuse~\cite{chhikara2025mem0, zhang2025gmemory, xu2025amem, langchain2026langmem}. However, these methods mainly support trajectory-level reuse rather than skill-level guidance. Retrieved memory records what happened before, but it may remain fragmented or trajectory-specific, rather than forming a persistent skill that specifies how the agent should act in future task executions. This motivates moving beyond retrieved memory toward skills that provide higher-level, persistent, and procedural guidance for future embodied task execution.

\subsection{Skills and Skill Self-Evolution for Agents}

This higher-level procedural guidance is commonly instantiated as skills. In robotics and physically grounded task settings, prior work often relies on reusable skills, action primitives, or programmatic policies to connect high-level instructions with executable actions. SayCan selects feasible robotic skills by combining language-model priors with affordance estimates~\cite{ichter2023saycan}; Code as Policies generates executable policy code by composing perception outputs and control primitives~\cite{liang2023codeaspolicies}; and ProgPrompt represents household task plans as program-like structures for situated execution~\cite{singh2023progprompt}. These methods demonstrate the value of skills for embodied task execution, but they mainly focus on selecting, composing, or executing existing skills rather than revising the skill itself from trajectories.

Recent work further studies skill self-evolution and procedural-memory evolution, where agents extract, refine, and reuse skills or reusable procedural knowledge from trajectories and interaction experience. Voyager builds an executable skill library through exploration in an interactive digital world~\cite{wang2024voyager}. Trace2Skill distills trajectory-local lessons into transferable agent skills through large-scale trajectory analysis and hierarchical consolidation~\cite{ni2026trace2skill}, while EvoSkills studies self-evolving multi-file skill packages through co-evolutionary verification~\cite{zhang2026coevoskills}. Other recent work explores related directions such as recursive skill construction, experience-derived executable skills, memory-skill evolution, procedural-memory refinement, and multimodal skill accumulation~\cite{xia2026skillrl,mi2026skillpro,zhang2026memskill,fang2026memp,cao2026reme,jiang2026xskill}. These works show that self-evolving skills are a promising paradigm for long-term agent improvement.

However, most existing skill self-evolution methods are not designed for physically grounded embodied environments, where trajectory outcomes are affected by perception, spatial grounding, object states, action preconditions, and execution reliability. As a result, coarse whole-skill update strategies may add redundant content, overwrite valid skill content, or preserve incorrect skill content. EmbodiSkill addresses this gap by making skill self-evolution skill-aware: each trajectory is used to produce targeted revision signals, and justified revisions are integrated through a skill-aware evolution spiral rather than rewriting the skill as a whole.

\section{Method}

\begin{figure}[h]
    \centering
    \includegraphics[
      width=0.95\linewidth,
      height=0.42\textheight,
      keepaspectratio
    ]{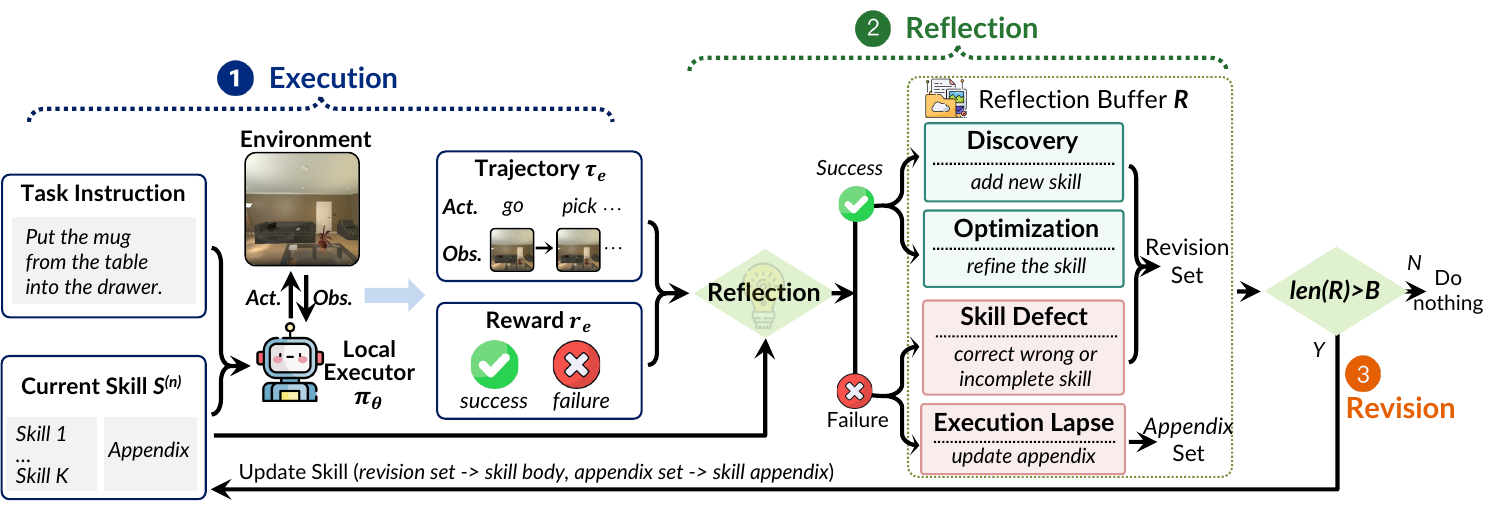}
    \caption{Overview of EmbodiSkill. The executor uses the current skill to perform embodied tasks and generate trajectories. Skill-aware reflection uses each trajectory to produce targeted reflection records. Accumulated reflections are consolidated into body-level revisions and skill-appendix updates, forming the next skill version. The revised skill then guides subsequent task execution, creating a Skill-Aware Evolution Spiral.}
    \label{fig2}
\end{figure}

\subsection{Problem Formulation}
\label{sec:problem_formulation}

\paragraph{Embodied trajectory.}
We consider an embodied agent interacting with a physical or physically grounded environment under natural-language task instructions. For a task instruction $I$, the agent observes the environment and produces actions over a finite horizon. We denote the resulting trajectory as
\begin{equation}
\tau = (I, o_1, a_1, \ldots, o_T, a_T, r),
\end{equation}
where $o_t$ is the observation at step $t$, $a_t$ is the action, $T$ is the trajectory length, and $r \in \{0,1\}$ is the final task success signal.

\paragraph{Skill-guided execution.}
At evolution step $n$, the agent is guided by a skill $S^{(n)}$, a persistent and revisable procedural specification for embodied task execution. We represent the skill as
\begin{equation}
S^{(n)} = \left(S_{\mathrm{body}}^{(n)}, S_{\mathrm{app}}^{(n)}\right),
\end{equation}
where $S_{\mathrm{body}}^{(n)}$ contains the main prescriptive skill content for future task execution, and $S_{\mathrm{app}}^{(n)}$ is a skill appendix that emphasizes valid content in $S_{\mathrm{body}}^{(n)}$. The appendix does not introduce new skill rules; it only highlights existing skill content that should be followed carefully during execution.

Given the current skill, the executor $\pi_\theta$ generates actions according to
\begin{equation}
a_t \sim \pi_\theta(\cdot \mid I, S^{(n)}, h_t),
\end{equation}
where $h_t = (o_1, a_1, \ldots, o_t)$ denotes the within-trajectory history. The executor parameters $\theta$ are kept fixed; all improvement across task executions is externalized into the evolving skill.

\paragraph{Skill evolution model.}
The executor $\pi_\theta$ uses the current skill to perform embodied tasks and produce trajectories. In contrast, the skill evolution model $F$ operates after trajectories are completed. We use $F$ as a unified notation for the model calls used in skill-aware reflection, reflection consolidation, skill-body revision, and skill-appendix update; these calls share the same underlying model but use different prompts and output formats.

\paragraph{Skill self-evolution objective.}
The goal of embodied skill self-evolution is to construct a sequence of skills
\begin{equation}
S^{(0)} \rightarrow S^{(1)} \rightarrow \cdots \rightarrow S^{(N)}
\end{equation}
such that later skills lead to higher task success when used by the same executor. Formally, we aim to improve
\begin{equation}
J(S; \pi_\theta) =
\mathbb{E}_{I,\mathcal{E}}
\left[
r(\tau)
\right],
\quad
\tau = \mathrm{Execute}(\pi_\theta, I, \mathcal{E}, S),
\end{equation}
where $\mathcal{E}$ denotes the embodied environment distribution.

\subsection{EmbodiSkill}
\label{sec:embodiskill}

EmbodiSkill implements embodied skill self-evolution as a skill-aware reflection and revision loop, as shown in Figure~\ref{fig2} and summarized in Algorithm~\ref{alg:embodiskill}. At evolution step $n$, the executor uses the current skill $S^{(n)}$ to perform embodied tasks, and each task execution produces a trajectory. After each trajectory, a skill evolution model uses the resulting trajectory to examine $S^{(n)}$ and produces up to $K$ targeted reflections, or no reflection when the trajectory does not provide reliable evidence for skill revision. Each valid reflection specifies the trajectory evidence, the corresponding revision directive, and either the implicated existing skill content or newly proposed skill content.

After a fixed number of valid reflections are accumulated, EmbodiSkill consolidates them before revising the skill. Reflections that propose body-level updates are consolidated to remove redundancy and resolve conflicts before updating $S_{\mathrm{body}}$. Reflections attributed to execution lapses do not revise the skill body; instead, after $S_{\mathrm{body}}$ is updated, they are used together with the updated skill body and the previous skill appendix to produce a new $S_{\mathrm{app}}$. The final revision step therefore yields the next skill version $S^{(n+1)}=(S_{\mathrm{body}}^{(n+1)}, S_{\mathrm{app}}^{(n+1)})$, where substantive changes to the skill body are grounded in skill-aware revision signals, while the skill appendix highlights valid skill content that should be followed carefully during later execution. The revised skill then guides subsequent task execution, forming a Skill-Aware Evolution Spiral.

\begin{algorithm}[t]
\caption{EmbodiSkill}
\label{alg:embodiskill}

\KwIn{Initial skill $S^{(0)}=(S_{\mathrm{body}}^{(0)}, S_{\mathrm{app}}^{(0)})$, executor $\pi_\theta$, skill evolution model $F$, task stream $\mathcal{I}$, revision interval $B$, maximum reflections per trajectory $K$}
\KwOut{Evolved skill $S$}

$(S_{\mathrm{body}}, S_{\mathrm{app}}) \leftarrow S^{(0)}$\;
$S \leftarrow (S_{\mathrm{body}}, S_{\mathrm{app}})$\;
$\mathcal{R} \leftarrow \emptyset$\tcp*[r]{reflection buffer}

\ForEach{task instruction $I \in \mathcal{I}$}{
    $\tau \leftarrow \textsc{Execute}(\pi_\theta, I, S)$\;
    
    $\mathcal{R}_{\tau} \leftarrow \textsc{SkillAwareReflect}(F, \tau, S, K)$\;
    
    \If{$\mathcal{R}_{\tau} \neq \emptyset$}{
        $\mathcal{R} \leftarrow \mathcal{R} \cup \mathcal{R}_{\tau}$\;
    }
    
    \If{$|\mathcal{R}| \ge B$}{
        $(\mathcal{R}_{\mathrm{disc}}, \mathcal{R}_{\mathrm{opt}}, \mathcal{R}_{\mathrm{def}}, \mathcal{R}_{\mathrm{lap}})
        \leftarrow \textsc{PartitionByType}(\mathcal{R})$\;
        
        $\widetilde{\mathcal{R}}_{\mathrm{rev}} \leftarrow
        \textsc{ConsolidateRevisions}(F, S_{\mathrm{body}},
        \mathcal{R}_{\mathrm{disc}}, \mathcal{R}_{\mathrm{opt}}, \mathcal{R}_{\mathrm{def}})$\;
        
        $S_{\mathrm{body}} \leftarrow
        \textsc{ReviseSkillBody}(F, S_{\mathrm{body}}, \widetilde{\mathcal{R}}_{\mathrm{rev}})$\tcp*[r]{targeted body edits}
        
        $S_{\mathrm{app}} \leftarrow
        \textsc{UpdateSkillAppendix}(F, S_{\mathrm{body}}, S_{\mathrm{app}}, \mathcal{R}_{\mathrm{lap}})$\tcp*[r]{update appendix only}
        
        $S \leftarrow (S_{\mathrm{body}}, S_{\mathrm{app}})$\;
        $\mathcal{R} \leftarrow \emptyset$\;
    }
}

\Return{$S$}\;

\end{algorithm}

\subsubsection{Skill-Aware Reflection}
\label{sec:skill_aware_reflection}

Skill-aware reflection determines how a trajectory should be interpreted with respect to the current skill. Given a trajectory $\tau$ and the current skill $S^{(n)}=(S_{\mathrm{body}}^{(n)}, S_{\mathrm{app}}^{(n)})$, the skill evolution model produces a set of reflection records:
\begin{equation}
\mathcal{R}_{\tau}
=
F(\tau, S^{(n)}, K)
=
\{\rho_i\}_{i=1}^{m_\tau},
\quad 0 \leq m_\tau \leq K,
\end{equation}
where $K$ is the maximum number of reflections allowed for one trajectory. When the trajectory does not provide reliable evidence for skill revision, $m_\tau=0$ and no reflection is produced.

Each reflection record $\rho_i$ contains a reflection type $c_i$, trajectory evidence $e_i$, and an update directive $d_i$. For reflection types that modify or emphasize existing skill content, the record also contains a target skill content $b_i$ that refers to a specific part of $S_{\mathrm{body}}^{(n)}$. The reflection type is chosen according to the trajectory outcome. For a successful trajectory, the reflection type is chosen from
\begin{subequations}
\begin{align}
c_i &\in
\{\textsc{Discovery}, \textsc{Optimization}\}, \quad r=1,
\end{align}
whereas for a failed trajectory, the reflection type is chosen from
\begin{align}
c_i &\in
\{\textsc{SkillDefect}, \textsc{ExecutionLapse}\}, \quad r=0.
\end{align}
\end{subequations}

\paragraph{DISCOVERY.}
A \textsc{Discovery} reflection indicates that the trajectory reveals useful skill content not covered by the current skill body. It does not require a target skill content, because it proposes new skill content rather than modifying existing content. Its directive $d_i$ specifies the new skill content to be considered during revision.

\paragraph{OPTIMIZATION.}
An \textsc{Optimization} reflection indicates that an existing target skill content is valid, but the trajectory suggests a more effective way to perform it. It must identify a target skill content $b_i$ and provide a revised version or modification directive for that target skill content.

\paragraph{SKILL DEFECT.}
A \textsc{SkillDefect} reflection indicates that an existing target skill content is incorrect, incomplete, or underspecified. It must identify the problematic target skill content $b_i$, provide trajectory evidence for the defect, and propose corrected skill content.

\paragraph{EXECUTION LAPSE.}
An \textsc{ExecutionLapse} reflection indicates that the target skill content is valid, but the executor fails to follow it in the trajectory. It must identify the valid target skill content $b_i$ and describe the deviation between what the skill requires and what the executor actually did. Its directive $d_i$ is not a body-level revision; instead, it produces appendix content for $S_{\mathrm{app}}$, reminding the executor to follow the corresponding valid skill content carefully during later task execution.

Thus, skill-aware reflection differs from trajectory summarization. It does not convert a trajectory into general feedback for rewriting the whole skill. Instead, it uses the trajectory to determine whether the trajectory supports adding new skill content, optimizing or correcting existing skill content, or updating the skill appendix, and produces structured signals for the revision procedure described next.

\subsubsection{Skill Revision}
\label{sec:skill_revision}

Skill revision converts accumulated skill-aware reflections into the next skill version. Once the reflection buffer reaches the revision interval $B$, EmbodiSkill partitions the buffered reflections by type:
\begin{equation}
(\mathcal{R}_{\mathrm{disc}}, \mathcal{R}_{\mathrm{opt}}, \mathcal{R}_{\mathrm{def}}, \mathcal{R}_{\mathrm{lap}})
=
\textsc{PartitionByType}(\mathcal{R}),
\end{equation}
where $\mathcal{R}_{\mathrm{disc}}$, $\mathcal{R}_{\mathrm{opt}}$, $\mathcal{R}_{\mathrm{def}}$, and $\mathcal{R}_{\mathrm{lap}}$ correspond to \textsc{Discovery}, \textsc{Optimization}, \textsc{SkillDefect}, and \textsc{ExecutionLapse}, respectively.

\paragraph{Consolidating body-level revision signals.}
The skill body is revised only from \textsc{Discovery}, \textsc{Optimization}, and \textsc{SkillDefect} reflections. Before editing the skill body, EmbodiSkill consolidates these reflections:
\begin{equation}
\widetilde{\mathcal{R}}_{\mathrm{rev}}
=
F
\left(
S_{\mathrm{body}}^{(n)},
\mathcal{R}_{\mathrm{disc}},
\mathcal{R}_{\mathrm{opt}},
\mathcal{R}_{\mathrm{def}}
\right).
\end{equation}
This consolidation step turns multiple local reflections into a consistent set of revision signals. It removes redundant reflections, merges overlapping suggestions, groups target-specific modifications by their target skill contents, and resolves conflicts among signals when possible. If a signal is misplaced or conflicts cannot be reliably resolved, the model either reassigns it to a more appropriate revision type or discards it, rather than forcing uncertain changes into the skill.

\paragraph{Revising the skill body.}
The consolidated revision signals are then used to update the skill body:
\begin{equation}
S_{\mathrm{body}}^{(n+1)} = F \left( S_{\mathrm{body}}^{(n)}, \widetilde{\mathcal{R}}_{\mathrm{rev}} \right).
\end{equation}
This step uses the skill evolution model as a constrained editor rather than a free-form rewriter. The editor revises the skill body according to the consolidated revision set $\widetilde{\mathcal{R}}_{\mathrm{rev}}$: \textsc{Discovery} adds new skill content, while \textsc{Optimization} and \textsc{SkillDefect} modify their identified target skill contents. Skill content not implicated by $\widetilde{\mathcal{R}}_{\mathrm{rev}}$ is kept unchanged in substance. The editor may still perform limited consistency edits, such as removing redundancy, normalizing format, or resolving local conflicts among the affected skill content. Thus, the revision updates targeted skill content while avoiding coarse rewriting of the whole skill.

\paragraph{Updating the skill appendix.}
After the skill body has been revised, EmbodiSkill updates the appendix part using \textsc{ExecutionLapse} reflections:
\begin{equation}
S_{\mathrm{app}}^{(n+1)}
=
F
\left(
S_{\mathrm{body}}^{(n+1)},
S_{\mathrm{app}}^{(n)},
\mathcal{R}_{\mathrm{lap}}
\right).
\end{equation}
Unlike the body-level revision step, this update does not introduce, delete, or rewrite skill rules in $S_{\mathrm{body}}$. Instead, it organizes execution-lapse evidence into appendix content anchored to the updated skill body. Each appendix item highlights a valid skill that the executor failed to follow, so that later executions pay closer attention to that skill. The appendix update can merge duplicate appendix items, remove obsolete appendix items that no longer correspond to the updated skill body, and incorporate new execution-lapse reflections.

The next skill version is therefore
\begin{equation}
S^{(n+1)}
=
\left(
S_{\mathrm{body}}^{(n+1)},
S_{\mathrm{app}}^{(n+1)}
\right).
\end{equation}
This design separates body-level skill revision from skill-appendix update: \textsc{Discovery}, \textsc{Optimization}, and \textsc{SkillDefect} reflections can change the skill body, while \textsc{ExecutionLapse} reflections only affect the appendix part. As a result, EmbodiSkill updates the skill selectively while reducing the risk of corrupting valid skill content.

\subsubsection{Skill-Aware Evolution Spiral}
\label{sec:skill_aware_evolution_spiral}

The Skill-Aware Evolution Spiral describes how EmbodiSkill turns repeated skill-aware revisions into progressive skill improvement. At evolution step $n$, the executor uses the current skill
$S^{(n)}=(S_{\mathrm{body}}^{(n)}, S_{\mathrm{app}}^{(n)})$
to perform embodied tasks. Each task execution produces a trajectory, which is reflected against the current skill to generate revision signals. These signals update the skill into
$S^{(n+1)}=(S_{\mathrm{body}}^{(n+1)}, S_{\mathrm{app}}^{(n+1)})$.
The revised skill is then used for subsequent task execution, producing new trajectories that may expose missing skill content, better execution strategies, skill defects, or execution lapses. In this way, the skill shapes future task execution, and the resulting trajectories provide new evidence for further skill evolution.

The spiral is driven by the interaction between the evolving skill and the trajectories it produces. When $S_{\mathrm{body}}$ is expanded, optimized, or corrected, the executor may handle task situations that were previously difficult, producing trajectories that expose new opportunities for further skill improvement. When $S_{\mathrm{app}}$ is updated from execution lapses, the executor is reminded to follow valid skill content more carefully during subsequent task execution. Across evolution steps, the skill therefore accumulates targeted improvements: $S_{\mathrm{body}}$ becomes more complete and accurate, while $S_{\mathrm{app}}$ makes valid skill content more salient during execution, improving the executability of the overall skill.

\section{Experiments}

\begin{table}[h]
\centering
\caption{
Task success rate (\%) on ALFWorld. We report overall success and six ALFWorld subtask categories. 
Bold indicates the best result in each column; underline indicates the second-best result on the overall score.
}
\label{tab:alfworld_results}
\small
\setlength{\tabcolsep}{4pt}
\begin{tabular}{l c cccccc}
\toprule
\multirow{2.5}{*}{\textbf{Method}} 
& \multirow{2.5}{*}{\textbf{Overall}} 
& \multicolumn{6}{c}{\textbf{ALFWorld Subtasks}} \\
\cmidrule(lr){3-8}
& & \textbf{Put} & \textbf{Clean} & \textbf{Heat} & \textbf{Cool} & \textbf{Examine} & \textbf{Puttwo} \\ 
\midrule

\multicolumn{8}{l}{\textbf{Closed-source direct agents}} \\
GPT-5.2        & 70.89 & 87.50 & 67.74 & 56.52 & 76.19 & 83.33 & 52.94 \\
Gemini-3-flash & 82.09 & 91.67 & 83.87 & 65.22 & 85.71 & 83.33 & 82.35 \\

\midrule
\multicolumn{8}{l}{\textbf{Executor: Qwen2.5-14B-Instruct}} \\
\multicolumn{8}{l}{\textit{Memory-based baselines}} \\
\quad No memory & 46.27 & 33.33 & 51.61 & 26.09 & 85.71 & 72.22 & 5.88 \\
\quad Mem0      & 12.69 & 0 & 0 & 4.35 & 0 & 88.89 & 0 \\
\quad G-Memory     & 67.16 & 50.00 & 83.87 & 69.57 & 85.71 & 83.33 & 17.65 \\
\quad LangMem   & 36.57 & 12.50 & 32.26 & 26.09 & 71.43 & 83.33 & 0.00 \\

\addlinespace[2pt]
\multicolumn{8}{l}{\textit{EmbodiSkill (Ours)}} \\
\quad EmbodiSkill w/ GPT-5.2        & 86.57 & 91.67 & \textbf{96.77} & 73.91 & \textbf{95.24} & 83.33 & 70.59 \\
\quad EmbodiSkill w/ Gemini-3-flash & 85.82 & 79.17 & 93.55 & 73.91 & 90.48 & \textbf{100.00} & 76.47 \\

\midrule
\multicolumn{8}{l}{\textbf{Executor: Qwen3.5-27B}} \\
\multicolumn{8}{l}{\textit{Memory-based baselines}} \\
\quad No memory & 61.19 & 66.67 & 51.61 & 65.22 & 76.19 & 72.22 & 35.29 \\
\quad Mem0      & 64.18 & 62.50 & 61.29 & 56.52 & 85.71 & 72.22 & 47.06 \\
\quad G-Memory     & 74.62 & 62.50 & 77.42 & \textbf{82.61} & 85.71 & 83.33 & 52.94 \\
\quad LangMem   & 62.69 & 62.50 & 61.29 & 52.17 & 76.19 & 72.22 & 52.94 \\

\addlinespace[2pt]
\multicolumn{8}{l}{\textit{EmbodiSkill (Ours)}} \\
\quad EmbodiSkill w/ GPT-5.2        & \textbf{93.28} & \textbf{95.83} & \textbf{96.77} & 73.91 & \textbf{95.24} & \textbf{100.00} & \textbf{100.00} \\
\quad EmbodiSkill w/ Gemini-3-flash & \underline{87.31} & \textbf{95.83} & 93.55 & 69.57 & \textbf{95.24} & 83.33 & 82.35 \\

\bottomrule
\end{tabular}
\end{table}

\begin{table*}[t]
\centering
\caption{
Task success rate (\%) on EmbodiedBench. Avg. reports the average success rate across subcategories, while subcategory scores are rounded to integers for readability.
Ours-GPT and Ours-Gemini denote EmbodiSkill using GPT-5.2 and Gemini-3-flash as the skill evolution model, respectively.
Bold indicates the best Avg. result for each benchmark; underline indicates the second-best Avg. result.
}
\label{tab:EmbodiedBench_results}
\small
\setlength{\tabcolsep}{2.3pt}
\renewcommand{\arraystretch}{1.05}
\begin{tabular}{lccccccc cccccc}
\toprule
\multirow{2.5}{*}{\textbf{Method}}
& \multirow{2.5}{*}{\textbf{Avg.}} 
& \multicolumn{6}{c}{\textbf{EB-Habitat}}
& \multirow{2.5}{*}{\textbf{Avg.}} 
& \multicolumn{5}{c}{\textbf{EB-Navigation}} \\
\cmidrule(lr){3-8} \cmidrule(lr){10-14}
& & \textbf{Base} & \textbf{Com.} & \textbf{Comp.} & \textbf{Vis.} & \textbf{Spa.} & \textbf{Long}
& & \textbf{Base} & \textbf{Com.} & \textbf{Comp.} & \textbf{Vis.} & \textbf{Long} \\
\midrule

\multicolumn{14}{l}{\textbf{Closed-source direct agents}} \\
GPT-5.2        & 40.00 & 80 & 40 & 26 & 44 & 32 & 18 & \underline{57.33} & 63 & 72 & 68 & 65 & 18 \\
Gemini-3-flash & 46.00 & 78 & 42 & 38 & 56 & 32 & 30 & 56.00 & 62 & 60 & 55 & 47 & 57 \\

\midrule
\multicolumn{14}{l}{\textbf{Executor: Qwen3-VL-8B-Instruct}} \\
\multicolumn{14}{l}{\textit{Memory-based baselines}} \\
\quad No memory & 24.00 & 58 & 20 & 20 & 18 & 18 & 10 & 45.00 & 58 & 53 & 62 & 50 & 2 \\
\quad Mem0      & 20.33 & 60 & 16 & 8  & 16 & 14 & 8  & 46.33 & 58 & 57 & 62 & 55 & 0 \\
\quad G-Memory     & 25.66 & 58 & 14 & 12 & 34 & 30 & 6  & 37.33 & 62 & 43 & 43 & 38 & 0 \\
\quad LangMem   & 19.67 & 60 & 12 & 8  & 16 & 14 & 8  & 47.00 & 60 & 57 & 62 & 57 & 0 \\

\addlinespace[2pt]
\multicolumn{14}{l}{\textit{EmbodiSkill (Ours)}} \\
\quad Ours-GPT    & 45.33 & 76 & 46 & 36 & 52 & 38 & 24 & 50.33 & 68 & 65 & 62 & 57 & 0 \\
\quad Ours-Gemini & 41.00 & 78 & 34 & 32 & 52 & 34 & 16 & 49.00 & 60 & 63 & 67 & 52 & 3 \\

\midrule
\multicolumn{14}{l}{\textbf{Executor: Qwen3-VL-32B-Instruct}} \\
\multicolumn{14}{l}{\textit{Memory-based baselines}} \\
\quad No memory & 34.67 & 78 & 26 & 28 & 34 & 28 & 14 & 46.33 & 60 & 53 & 60 & 53 & 5 \\
\quad Mem0      & 38.33 & 84 & 24 & 26 & 42 & 34 & 20 & 52.00 & 62 & 60 & 65 & 67 & 7 \\
\quad G-Memory     & 45.00 & 92 & 24 & 36 & 50 & 32 & 34 & 50.33 & 52 & 53 & 58 & 65 & 23 \\
\quad LangMem   & 38.33 & 86 & 26 & 28 & 40 & 32 & 18 & 52.00 & 62 & 60 & 65 & 67 & 7 \\

\addlinespace[2pt]
\multicolumn{14}{l}{\textit{EmbodiSkill (Ours)}} \\
\quad Ours-GPT    & \underline{50.33} & 92 & 46 & 50 & 54 & 34 & 26 & \textbf{61.33} & 65 & 68 & 73 & 67 & 33 \\
\quad Ours-Gemini & \textbf{52.33} & 96 & 44 & 52 & 62 & 36 & 24 & \textbf{61.33} & 67 & 70 & 70 & 68 & 32 \\

\bottomrule
\end{tabular}
\end{table*}

\subsection{Experimental Setup}
\label{sec:experimental_setup}

\paragraph{Benchmarks and metrics.}
We evaluate EmbodiSkill on three embodied task benchmarks: ALFWorld~\cite{shridhar2021alfworld}, EmbodiedBench-Habitat, and EmbodiedBench-Navigation~\cite{yang2025embodiedbench}. ALFWorld contains household task completion problems in interactive environments, involving object search, navigation, container interaction, and object state changes. EmbodiedBench-Habitat evaluates visual object interaction in 3D environments, while EmbodiedBench-Navigation evaluates visual navigation. In our split, ALFWorld contains 3,553 training tasks and 134 test tasks; EmbodiedBench-Habitat contains 1,000 training tasks and six test subsets with 50 tasks each; and EmbodiedBench-Navigation contains 1,000 training tasks and five test subsets with 60 tasks each. We report task success rate for all benchmarks.

\paragraph{Skill evolution protocol.}
EmbodiSkill evolves skills from trajectories collected on training tasks and evaluates the evolved skill on held-out test tasks. We set the maximum number of reflections per trajectory to $K=1$ and perform 10 skill revision stages unless otherwise specified. Training tasks are sampled in randomized order, and the training stream is reshuffled when additional trajectories are required. During test evaluation, the evolved skill is fixed.

\paragraph{Models.}
For ALFWorld, we instantiate the executor $\pi_\theta$ with Qwen2.5-14B-Instruct~\cite{qwen2024qwen25} and Qwen3.5-27B~\cite{qwen2026qwen35}. For EmbodiedBench-Habitat and EmbodiedBench-Navigation, Qwen3-VL-8B-Instruct and Qwen3-VL-32B-Instruct as executors~\cite{qwen2025qwen3vl}. The skill evolution model $F$ is instantiated with GPT-5.2~\cite{openai2025gpt52} or Gemini-3-flash~\cite{google2025gemini3flash}. All executor parameters are kept fixed during skill evolution, so performance gains come from the evolving skill rather than model parameter updates.

\paragraph{Baselines.}
We compare EmbodiSkill with two groups of baselines in the main results. First, closed-source direct agents use GPT-5.2 or Gemini-3-flash directly for task execution without an evolving skill. Second, memory-based methods, including Mem0~\cite{chhikara2025mem0}, G-Memory~\cite{zhang2025gmemory}, and LangMem~\cite{langchain2026langmem}, store and retrieve trajectory-level information to guide task execution. For methods using the same local executor, the executor is kept identical so that performance differences reflect the external skill or memory mechanism rather than changes in the executor.

\subsection{Main Results}
\label{sec:main_results}

Table~\ref{tab:alfworld_results} and Table~\ref{tab:EmbodiedBench_results} report the main results on ALFWorld and EmbodiedBench. Overall, EmbodiSkill improves task success by evolving an external skill from training trajectories, showing stronger performance than direct model execution and trajectory-level memory retrieval.

\paragraph{Results on ALFWorld.}
On ALFWorld, EmbodiSkill achieves the best overall performance among all evaluated methods. With Qwen3.5-27B as the executor and GPT-5.2 as the skill evolution model, EmbodiSkill reaches 93.28\% task success. This outperforms GPT-5.2 used as a direct agent by 31.58\% and Gemini-3-flash used as a direct agent by 13.63\%. Compared with memory-based methods under the same Qwen3.5-27B executor, EmbodiSkill exceeds the strongest memory baseline, G-Memory, by 25.01\%. These results indicate that consolidating training trajectories into an evolving skill provides stronger guidance than directly executing with a frontier model or retrieving trajectory-level memory.

The subtask breakdown further shows that the improvement is not concentrated in a single task type. EmbodiSkill with Qwen3.5-27B and GPT-5.2 achieves the best or tied-best performance on five of the six ALFWorld subtask categories, including Put, Clean, Cool, Examine, and Puttwo. The gain is especially clear on Puttwo, where EmbodiSkill reaches 100.00\%, compared with 52.94\% for G-Memory under the same executor. This suggests that the evolved skill is particularly helpful for multi-step household tasks that require object search, state tracking, and correct action ordering.

\paragraph{Results on EmbodiedBench.}
Table~\ref{tab:EmbodiedBench_results} reports results on visual embodied object interaction and navigation. On EB-Habitat, EmbodiSkill with Qwen3-VL-32B-Instruct and Gemini-3-flash achieves the best average score of 52.33\%, outperforming the strongest memory-based baseline by 16.29\% and the strongest closed-source direct agent by 13.76\%. On EB-Navigation, EmbodiSkill with Qwen3-VL-32B-Instruct reaches the best average score of 61.33\% with both GPT-5.2 and Gemini-3-flash as the skill evolution model, outperforming the strongest memory-based baseline by 17.94\% and the strongest closed-source direct agent by 6.98\%. These results show that EmbodiSkill also improves task success in visual embodied settings, where agents must rely on visual observation and spatial grounding to complete object interaction and navigation tasks.

\begin{table}[htbp]
\centering
\caption{
Ablation of skill self-evolution and skill-aware reflection on ALFWorld. We report task success rate (\%). 
$\Delta_{\mathrm{aware}}$ denotes the improvement of EmbodiSkill over skill-unaware evolution.
}
\label{tab:skill_aware_ablation}
\small 
\setlength{\tabcolsep}{4.5pt} 
\begin{tabular}{ll cccc c}
\toprule

\multirow{2.5}{*}{\textbf{Executor}} 
& \multirow{2.5}{*}{\textbf{Skill Model}} 
& \multicolumn{4}{c}{\textbf{Configurations}} 
& \multirow{2.5}{*}{$\boldsymbol{\Delta_{\mathrm{aware}}}$} \\
\cmidrule(lr){3-6}
& & \textbf{No skill} & \textbf{Static Skill} & \textbf{Skill-unaware} & \textbf{EmbodiSkill} & \\

\midrule
\multirow{2}{*}{Qwen2.5-14B} 
& GPT-5.2        & \multirow{2.3}{*}{46.27} & 65.67 & 70.90 & 86.57 & +15.67 \\
& Gemini-3-flash &                        & 58.95 & 67.91 & 85.82 & \textbf{+17.91} \\

\midrule
\multirow{2}{*}{Qwen3.5-27B} 
& GPT-5.2        & \multirow{2.3}{*}{61.19} & 73.13 & 78.36 & \textbf{93.28} & +14.92 \\
& Gemini-3-flash &                        & 79.85 & 85.82 & 87.31 & +1.49 \\

\bottomrule
\end{tabular}
\end{table}

\begin{figure}[t]
    \centering
    \includegraphics[width=0.78\linewidth]{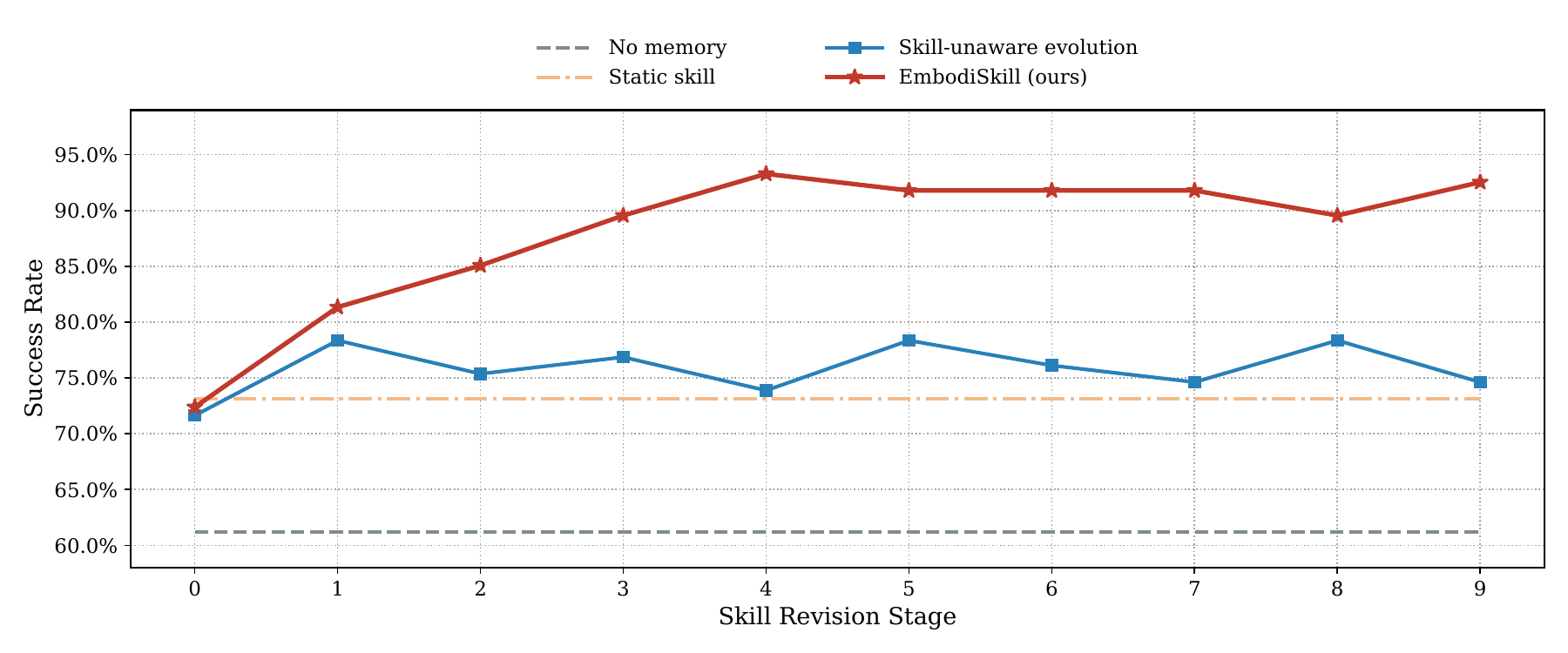}
    \caption{
    ALFWorld test success rate across skill revision stages. EmbodiSkill quickly improves from the static skill and remains at a high success rate, while skill-unaware evolution converges to a lower performance range.
    }
    \label{fig:skill_evolution_curve}
\end{figure}

\subsection{Ablation and Analysis}
\label{sec:ablation_analysis}

\paragraph{Ablation of skill self-evolution and skill-aware reflection.}
To analyze where the improvement of EmbodiSkill comes from, we compare four settings on ALFWorld. \textit{No memory} directly uses the executor without external memory or skill evolution. \textit{Static Skill} uses the initial skill without further revision. \textit{Skill-unaware} updates the skill from trajectories, but revises the skill coarsely without explicitly identifying the implicated skill content, revision type, or separation between skill-body revision and skill-appendix update. EmbodiSkill performs the full skill-aware reflection and revision process.

As shown in Table~\ref{tab:skill_aware_ablation}, both skill self-evolution and skill-aware reflection contribute to performance. With Qwen3.5-27B as the executor and GPT-5.2 as the skill evolution model, the static skill improves over the no-memory setting from 61.19\% to 73.13\%, corresponding to a 19.51\% relative improvement and showing the benefit of skill-guided execution. Skill-unaware evolution further improves the result to 78.36\%, but remains substantially below EmbodiSkill, which reaches 93.28\%. This corresponds to a 27.55\% relative improvement over the static skill and a 19.04\% relative improvement over skill-unaware evolution. Similar gains are observed with Qwen2.5-14B-Instruct, where EmbodiSkill improves over skill-unaware evolution by 22.10\% with GPT-5.2 and 26.37\% with Gemini-3-flash. These results show that the gain does not come only from updating the skill, but from making the update skill-aware and targeted.

\paragraph{Evolution over skill revision stages.}
We further examine how performance changes during skill self-evolution. Figure~\ref{fig:skill_evolution_curve} shows the ALFWorld test success rate across 10 skill revision stages under the Qwen3.5-27B executor and GPT-5.2 skill evolution model. EmbodiSkill rapidly improves from the static skill performance of 73.13\% and reaches 93.28\% during evolution, after which it remains in a high performance range. In contrast, skill-unaware evolution improves over the static skill but converges to a lower performance range and exhibits larger fluctuations. This suggests that directly revising the skill without skill-aware attribution can make evolution less stable, while EmbodiSkill produces more reliable improvement by updating targeted skill content.

\section{Conclusion}
\label{sec:conclusion}

We presented EmbodiSkill, a framework for embodied skill self-evolution. EmbodiSkill uses skill-aware reflection to examine trajectories with respect to the current skill and produce targeted revision signals, avoiding coarse whole-skill rewriting. These signals are integrated through a Skill-Aware Evolution Spiral, where revised skills guide subsequent task execution and new trajectories provide further evidence for skill improvement. Experiments on ALFWorld and EmbodiedBench show that EmbodiSkill improves task success over direct agents and memory-based baselines, and ablations confirm the importance of skill-aware revision. These results suggest that skill-aware self-evolution is a promising way to build embodied agents that accumulate reusable procedural knowledge from their own trajectories.

{
\small
\bibliographystyle{unsrt}
\bibliography{thuair}
}







\end{document}